# Utilizing Large Language Models to Generate Synthetic Data to Increase the Performance of BERT-Based Neural Networks

Chancellor R. Woolsey[1], Prakash Bisht[1], Joshua Rothman[2], MD, MS, Gondy Leroy[1], PhD, MS

[1]University of Arizona, Tucson, Arizona, [2]University of California San Diego, San Diego, California

**Abstract**

*An important issue impacting healthcare is a lack of available experts. Machine learning (ML) models could resolve this by aiding in diagnosing patients. However, creating datasets large enough to train these models is expensive. We evaluated large language models (LLMs) for data creation. Using Autism Spectrum Disorders (ASD), we prompted ChatGPT and GPT-Premium to generate 4,200 synthetic observations to augment existing medical data. Our goal is to label behaviors corresponding to autism criteria and improve model accuracy with synthetic training data. We used a BERT classifier pre-trained on biomedical literature to assess differences in performance between models. A random sample (N=140) from the LLM-generated data was evaluated by a clinician and found to contain 83% correct example-label pairs. Augmenting data increased recall by 13% but decreased precision by 16%, correlating with higher quality and lower accuracy across pairs. Future work will analyze how different synthetic data traits affect ML outcomes.*

**Introduction**

Timely screening and diagnosis can be a problem for many medical conditions[1,2,3]. The usage of machine learning can help. New deep-learning models that exhibit excellent performance in classifying doctors' notes on medical records may be especially helpful. For example, Dong et al.[4] created a system for classifying Chinese electronic medical records. However, these deep-learning ML models require large amounts of data to perform satisfactorily. Such data is not always available to researchers, or it is sometimes available from data brokers but is very expensive. An alternative approach may be the creation of artificial data.

Different types of data are used in machine learning, ranging from structured data to sounds, images, and text. With the increasing availability of generative AI, large language models (LLMs) may be especially suitable for creating otherwise difficult-to-mimic data such as images or text. We focus on text in our project in the context of improving early screening and diagnosis of autism spectrum disorders (ASD). Autism is a mental health disorder that can be detected early on in an individual's life and impacts much of the population, with an estimated 1.56% of all children receiving this diagnosis by the age of four[5] and nearly 2.78% of all children by the age of eight[6]. It is demarcated by individuals who struggle in social interactions, communication, and repetitive actions or activities as clinically defined in the Diagnostic and Statistical Manual of Mental Disorders[7]. An autism diagnosis is based on showing observable, continuous setbacks in "social communication and interaction" (labeled as 'A' symptoms in the DSM) as well as "restricted, repetitive behaviors" (labeled as 'B' symptoms in the DSM).

We are working on deep learning machine learning models to detect ASD using information in electronic health records (EHR). We work with free text in EHR records reviewed and labeled by experts. Each sentence in the text received a label based on DSM diagnostic criteria or no label. However, since such expert review is time-consuming and expensive, the datasets available for training and testing are relatively small. We are evaluating LLMs, i.e., different versions of ChatGPT, for their potential to supply synthetic data for the ML. We evaluate the effect of adding synthetic data on BioBERT, a BERT model pre-trained on thousands of biomedical abstracts[8].

**Related Work**

*Generating Artificial Data to Improve Machine Learning*
Different types of data can be created artificially. For example, for image data such as x-rays or photographs, Yoshihiro Shima[9] applied a series of rotations, translations, skews, and distortions to increase accuracy by over 10.00% in classifying images of airplanes, birds, cars, cats, deer, dogs, horses, monkeys, ships and trucks. Nagaraju et al.[10] applied addition using scaling, flips, blurs, and color changes to achieve a boost inaccuracy in the classification of maize crop disease by more than 5.00%.

However, techniques appropriate for images are not optimal for creating text-based data (i.e., medical observations, patient comments, etc.). Ahlbäck and Dougly[11] focused on taking their current datasets and introducing spelling errors, word insertions from embeddings, or greyscaling (replacing "strong" words with less extreme synonyms, i.e., changing "extremely" to "quite") and were able to increase F1 from of 0.945 to 0.966 on fake news detection.

More advanced approaches have turned to generative AI technologies to artificially increase the quantity of text data through LLMs. Many of these studies focus on prompting an LLM with some conditions to get the desired results. To help with this, many of these studies provide example data as a starting point. Sahu et al.[12] prompted GPT-3 with the task of sentence completion and provided a set of ten examples to study intent classification. They were able to boost accuracy by 5% across four different datasets ranging from banking to multi-purpose data. Dai et al.[13] used GPT-3 to generate six new sentences for text classification based on a single sentence from a dataset. They found that this increases the accuracy of their BERT-based networks by an average of 0.1315 from raw datasets from Amazon, a set of transcribed medical symptoms, and a set containing 20,000 different biomedical abstracts.

Other models have attempted to prompt LLMs without providing starting examples with promising results. Timo Schick and Hinrich Schütze[14] used GPT-2 and developed the DINO framework to help users generate datasets from pre-trained models via prompts to achieve an average semantic textual similarity (STS) score of 75.20 across a variety of STS benchmarks – outperforming second place by almost one point. This pattern recognition capability was then explored further by Li et al.[15], who incorporated Schick and Schütze's DINO framework for text generation, but this time added the requirement that each response maintain the usage of specific vocabulary terms. This process achieved a boosted 6 points higher ROUGE F1/Precision/Recall of 36.27/16.78/33.83 compared to a baseline of 29.86/10.19/27.08 FlanT5 (finetuned language model) from Chung et al.'s[16] previous work.

*Problems with Artificial Data*
LLMs are imperfect. Potential bias is a shortcoming that can affect outcomes. Salinas et al.[17] found biases against individuals of Mexican nationality based on a series of prompts asking for job recommendations. They found that Mexicans were recommended roles such as "Engineer; Software; Quality; Assurance" more than six times less often than other nationalities and found that Mexicans were predicted to have the smallest median salary out of any nationality. Additionally, Nicole Gross[18] found that ChatGPT still shows gender bias in its responses, noting that it perpetuates gender-stereotyping roles of mothers and fathers within family units and even ranks the value of technical skills and communications skills differently in value for CVs depending on if the "author" of the CV was a woman or a man.

In medicine, if an LLM maintains such biases against individuals of different backgrounds, it can increase the risk of a wrong diagnosis or not offering the correct treatment based on preconceived notions of gender, nationality[19], or other potential biases. If these problems persist into data generation, it may not yield reliable results that can be trained on.

**Methods**
*Data Set and Objective*
As stated above, autism is diagnosed through the identification of problematic behaviors that are labeled as A1-A3 and B1-B4 diagnostic criteria as described in DSM5[7]. The descriptions and examples of the criteria can be seen below (Table 1).

Our dataset is part of the CDC surveillance data from the state of Arizona, and the free text is labeled by trained experts. This dataset is highly unbalanced, and only 14.30% of sentences within our medical observation dataset contain a diagnostic label. Additionally, out of our entire usable dataset, fewer than 300 examples correspond with B3 symptoms, and more than 1,000 correspond to an A1 symptom. This low number of examples leads to lower performance of ML algorithms[20,21].

**Table 1:** A list of autism labels corresponding to symptoms and examples used by the DSM5.

| Label | Symptom | Example |
|---|---|---|
| A1 | Lack of emotional reflection. | Struggles in maintaining back-and-forth dialogue |
| A2 | Struggles in using nonverbal communication. | Poor/abnormal eye contact, lack of understanding of body language |
| A3 | Issues in creating, keeping, and grasping how to handle relationships. | Maladjustment of behavior to match differing social environments |
| B1 | Common, pattern-like behaviors in relation to speaking, moving, or using items. | Abnormal repeated usage of certain phrases, specific ordering of toys |
| B2 | Observation of specific behaviors with issues if such rituals are not followed. | Distress if a specific food is not eaten at the same time every day |
| B3 | Hyperfixation on interests differing from common norms. | Interests in the behaviors of fans or curtains as they interact with light |
| B4 | Either major sensitivity or lack-thereof to sensory traits in a given space. | Pain-like reactions to certain sounds but little concern for extreme heat |

*Model Used*

We use a BioBERT multilabel neural network pretrained on biomedical abstracts and articles from PubMed. For our implementation, we used the "TFBertModel" from the HuggingFace transformers library[22] with an attention mask to equalize the input data[23]. After the mask and BioBERT layers, we use a 1-dimensional global average pooling and a dense layer for a total of 108,315,655 parameters.

*Artificial Data Generation*

There have been multiple iterations of OpenAI's GPT model before its official release as ChatGPT. For our study, we used OpenAI's GPT-3.5-Turbo and OpenAI's GPT-4 to generate data. GPT-3.5-Turbo is the most current version of 175-billion-parameter GPT-3[24,] and GPT-4 is only accessible via a "ChatGPT Plus" subscription[25]. However, OpenAI claims it is their "most capable model"[26].

Ideally, GPT-3.5 would have been studied independently, but we found that GPT-3.5 did not return more than 25 examples at a time, a shortcoming that was later resolved. We show one of our attempts to prompt GPT-3.5 to return more examples in Table 2.

To avoid these problems, we created data for two studies. First, we generated examples using both GPT-3.5 and GPT-4 and changed as few words as possible in each prompt. The viewpoint was kept constant: "an expert pediatrician with over ten years of experience in the field, and with that, you have become famous for your understanding of autism in young c," where 'c' represents the subject's gender (boy, girl, or 'child' for gender neutral cases).

However, our study changed three different elements in each prompt:
- The gender the 'doctor' was studying: male, female, neutral (child)
- The DSM5 symptom we wanted more examples of: A1, A2, A3, B1, B2, B3, or B4
- The number of children who were being studied: 25 or 150

For example, if we wanted a prompt that returned 25 examples of boys showing behaviors as described by A1, it would be like the following:
- You are an expert pediatrician with over ten years of experience in the field, and with that, you have become famous for your understanding of autism in young boys. Today, you have been presented 25 boys experiencing the following known autism symptom: "Deficits in social-emotional reciprocity, ranging, for example, from abnormal social approach and failure of normal back-and-forth conversation; to reduced sharing of interests, emotions, or affect; to failure to initiate or respond to social interactions." I need you to uniquely record each of the 25 boys from the doctor's point of view. If possible, please write this as though this would end up on the boys' permanent medical record and keep each observation limited to only 1-2 sentences each. Please note, these are medical observations, not perfect descriptions for each young boy. Thus, if short phrases make the observation clear, then do those. Please note: I need to see each observation for each little boy; there cannot be any ellipses to skip over any of the boys. Please number each one from 1 to 25. Thank you.

Using this setup, we requested 25 observations per symptom across three genders (boy, girl, and child) for a total of 1050 new lines of text data and 150 new lines per symptom.

**Table 2:** An example of a prompt that failed to generate data due to the number of responses asked for

| Prompt | Response |
|---|---|
| You are an expert pediatrician with over ten years of experience in the field, and with that, you have become famous for your understanding of autism in young boys. Today, you have been presented 30 boys experiencing the following known Autism symptom: "Deficits in social-emotional reciprocity, ranging, for example, from abnormal social approach and failure of normal back-and-forth conversation; to reduced sharing of interests, emotions, or affect; to failure to initiate or respond to social interactions." I need you to uniquely record each of the 30 boys from the doctor's point of view. If possible, please write this as though this would end up on the boys' permanent medical record and keep each observation limited to only 1-2 sentences each. Please note, these are medical observations, not perfect descriptions for each young boy. Thus, if short phrases make the observation clear, then do those. Please note: I need to see each observation for each little boy, there cannot be any ellipses to skip over any of the boys. Please number each one from 1 to 30. Thank you. | I'm sorry for any confusion, but it's important to clarify that I don't have access to real-time data or the ability to interact with specific individuals. I also cannot generate 30 unique observations without more specific information about each case. If you'd like, I can provide you with a general example of how these observations might look based on the information you've provided. Please let me know if you'd like me to proceed with that. |

Second, we generated examples using GPT-4. We requested 150 observations across the same three genders for a total of 3,150 responses and 450 new lines per symptom. Table 3 shows our count of baseline examples compared to the total number of examples, including the synthetic data for each label.

**Table 3:** A count of all examples in the original dataset, the dataset augmented with GPT-3.5 and GPT-4, and the dataset augmented with GPT-4 only.

| Label | Count of Examples per Labels | Examples augmented with GPT-3.5 and GPT-4 | Examples augmented with GPT-4 Only |
|---|---|---|---|
| A1 | 1079 | 1229 | 1529 |
| A2 | 580 | 730 | 1030 |
| A3 | 667 | 817 | 1117 |
| B1 | 652 | 802 | 1102 |
| B2 | 443 | 593 | 593 |
| B3 | 215 | 365 | 665 |
| B4 | 848 | 998 | 1298 |

*Evaluation*

First, we run the model using only existing EHR data and no synthetic data. This run serves two purposes: to provide a baseline to compare augmented datasets to (EHR plus synthetic data) and to provide the initial finetuning for our BioBERT models. Here, we also count the total samples of observations for each criterion to see what models might be impacted the most by our data augmentation. Next, we analyze sentence starts to get a preliminary look at potential differences between our original data and the synthetic data.

We also conducted a domain expert evaluation. Our expert is an assistant professor and pediatrician at UC San Diego Health with over four years of experience. We randomly selected 140 text observations from the GPT-3.5 and GPT-4 data. The clinical expert evaluated each observation on four dimensions using a 1-5 scale:
- the behavior is common in autistic individuals (1: rare – 5: common).
- the behavior is common in people in general (1: rare – 5: common)
- the behavior is normal for a different mental health ailment other than autism (e.g. ADHD, anxiety, etc.)(1: rare – 5: common)
- if the data could be found in real EHR notes (1: rare – 5: common)

Additionally, he also noted if the given label was correct, incorrect, or missing additional labels (e.g. A1 did not fully describe the behavior, it would also need an A2 label)

Finally, we train and test our BioBERT model using the original and augmented data sets. In each case, we used the data to fine-tune the BioBERT model, tested on a separate ADDM test set and reported precision, recall, and the F-measure. Since we have unbalanced data with the majority of sentences not receiving a label, i.e., the majority of sentences in an EHR are not descriptions of autistic behaviors, we do not report accuracy. Accuracy would be artificially high and does not contribute to understanding the results.

**Results**

*Data Set Quantification*

First, we counted the number of examples with similar sentence starts (first word only). Table 4 shows the top 10 similar examples, the starting word, and the observation frequency. We also provide the proportion of the dataset these sentences constitute, with the baseline data subset (comprising of only test with DSM5 criteria) consisting of 57.52% of the data, the GPT-3.5 and GPT-4 subset consisting of 75.33% of the new data, and the GPT-4 only subset consisting of 39.81% of the new data.

**Table 4:** Ten most frequent examples based on initial word similarity.

| Baseline (N=5351) | | GPT-3.5 and GPT-4 (N=1050) | | GPT-4 Only (N=3150) | |
|---|---|---|---|---|---|
| Start Word | Count (%) | Start Word | Count (%) | Start Word | Count (%) |
| He | 33.04 | Patient | 14.10 | Struggles | 5.81 |
| She | 8.26 | Demonstrate | 11.52 | Rare | 5.24 |
| His | 4.07 | Display | 9.43 | Rarely | 5.17 |
| Child | 2.56 | Child | 8.86 | Show | 4.57 |
| When | 1.96 | exhibit/Exhibit | 8.19 | Displays | 3.52 |
| In | 1.87 | Struggles | 5.71 | Does | 3.49 |
| The | 1.61 | engage/Engage | 5.52 | Avoid | 3.46 |
| Sensory | 1.59 | Shows | 5.33 | Repeat | 2.95 |
| At | 1.40 | Observe | 4.19 | Appear | 2.89 |
| During | 1.16 | does | 2.48 | Demonstrates | 2.70 |
| % of Dataset | 57.52 | % of Dataset | 75.33 | % of Dataset | 39.81 |

The expert evaluation of the synthetic is shown in Table 5. Overall, the generated observation was highly common for autistic individuals, highly uncommon for individuals without autism, and highly uncommon for people with a mental health issue other than autism and such behaviors would almost always be found in clinical notes if such behaviors were to be observed.

Additionally, Table 6 shows the percentage of observations that correctly corresponded to a given label. Out of all the observations, 83.00% of the labels matched the prompted label, 7.00% of the labels were incorrect, and 10.00% of the labels required extra labels to be correct.

**Table 5:** Expert scores for synthetic data

| ChatGPT-3.5 and ChatGPT-4 Scores | | | | | |
|---|---|---|---|---|---|
| Criterion Label | N | Typical for Autism | Is Normal | Behavior but not typical for autism | Would also be found in EHR clinical notes |
| A1 | 18 | 5 | 1 | 1 | 5 |
| A2 | 24 | 5 | 1.04 | 1.04 | 5 |
| A3 | 22 | 5 | 1.05 | 1 | 5 |
| B1 | 16 | 4.88 | 1.19 | 1.13 | 5 |
| B2 | 18 | 4.94 | 1.06 | 1.06 | 5 |
| B3 | 22 | 5 | 1.18 | 1.05 | 5 |
| B4 | 20 | 5 | 1.4 | 1.2 | 4.95 |
| Average | 20 | 4.98 | 1.13 | 1.06 | 4.99 |

**Table 6:** Percent of observations that have the correct label according to expert

| ChatGPT-3.5 and ChatGPT-4 Percent Correct | | | |
|---|---|---|---|
| Criterion Label | Total Correct (%) | Total Incorrect (%) | Total Incomplete (%) |
| A1 | 50.00 | 22.22 | 27.78 |
| A2 | 91.67 | 0 | 8.33 |
| A3 | 68 | 9.10 | 22.73 |
| B1 | 100 | 0 | 0 |
| B2 | 88.89 | 5.56 | 5.56 |
| B3 | 86.36 | 9.10 | 4.45 |
| B4 | 100 | 0 | 0 |
| Overall | 83.57 | 6.43 | 10.00 |

*Machine Learning Outcomes*

Table 7 shows the results for ML labeling using a BioBERT classifier fine-tuned on the baseline data, followed up by the augmented data (for both GPT-3.5 and GPT-4 and GPT-4 only). For GPT-3.5 and GPT-4, we found a 6% increase in recall from the baseline. For GPT-4 only, we found a 13% increase in the average recall. Between both sets, there was an increase in recall, with the only label that did not experience an increase in both being B4. For B4, the recall dropped by 3% in the combined GPT-3.5 and GPT-4 and 0.05 in the GPT-4 only data. However, recall also dropped by 5% in the GPT-3.5 and GPT-4 for the B1 symptom.

However, precision decreased. For GPT-3.5 and GPT-4, we found an average decrease of 4%. For GPT-4 only, we found a decrease of 0.16. The only two labels to experience an increase from the baseline were B1 and B4 from the combined GP-T3.5 and GPT-4. Out of all precision measures, the only time the GPT-4 extended set was not below the baseline was B4.

**Discussion and Conclusion**

**Table 7**: Performance of the Classification using the Baseline and Augmented Datasets

|  | Original Data Set | | | Data Set augmented with GPT-3.5 and GPT-4 | | | Data Set Augmented with GPT-4 | | |
|---|---|---|---|---|---|---|---|---|---|
| Label | Recall | Precision | F1 | Recall | Precision | F1 | Recall | Precision | F1 |
| A1 | 0.33 | 0.73 | 0.45 | 0.46 | 0.63 | 0.53 | 0.52 | 0.62 | 0.56 |
| A2 | 0.59 | 0.83 | 0.69 | 0.66 | 0.76 | 0.70 | 0.81 | 0.61 | 0.7 |
| A3 | 0.50 | 0.74 | 0.60 | 0.58 | 0.65 | 0.61 | 0.70 | 0.50 | 0.58 |
| B1 | 0.67 | 0.64 | 0.66 | 0.62 | 0.68 | 0.65 | 0.69 | 0.47 | 0.56 |
| B2 | 0.56 | 0.76 | 0.65 | 0.63 | 0.72 | 0.67 | 0.76 | 0.49 | 0.60 |
| B3 | 0.35 | 0.65 | 0.46 | 0.46 | 0.57 | 0.51 | 0.49 | 0.55 | 0.52 |
| B4 | 0.65 | 0.64 | 0.64 | 0.62 | 0.65 | 0.63 | 0.60 | 0.64 | 0.62 |
| Average | 0.52 | 0.71 | 0.59 | 0.58 | 0.67 | 0.62 | 0.65 | 0.55 | 0.59 |

Through the usage of ChatGPT, we augmented an unbalanced dataset of medical observations to include more examples of each criterion for autism to increase the performance of our classifiers. Based on an expert review of our data, ChatGPT was able to generate data that would be seen in someone with autism that, if combined with other behaviors, would more than likely lead to an autism diagnosis. Through the implementation of text data created by ChatGPT-3.5 and GPT-4, we were able to successfully increase the recall of our classifier. However, these results come at the cost of a drop in the classifier's precision.

Based on the results from Table 4, GPT-3.5 and GPT-4 tended to avoid starting the medical observations with pronouns, prepositions, articles, or adverbs and instead start sentences with active verbs or proper nouns. This might have come as a result of the instructions in the prompt, which our instructions to the GPT models state, "If short phrases make the observation clear, then do those." This "phrase-like" structure can be seen in Table 1, where each example used came from our synthetic data. However, this cannot be claimed with certainty. Table 4 only shows the lack of uniformity of starting words in our datasets.

While the medical expert leaves near-perfect reviews on whether a given observation would be considered a sign of autism, the labels themselves were only 100% correct (meaning no additional labels needed) 83.57% of the time. On a sample size of 140 medical observations, a miss-rate (meaning the total times the labels were incorrect) of 16.43% may be considered small. However, our two studies comprised 1,050 and 3,150 observations, respectively. This 16.43% miss-rate translates to 173 and 518 incorrect observations. Thus, with domain-specific ChatGPT-generated text data, an expert review of the data is needed to remove incorrect pieces.

Notably, the B1 criteria ranked the lowest for behaviors common in individuals with autism and highest for normal behavior for individuals without autism. This aligns with the lower recall compared to the baseline on the combined GPT-3.5 and GPT-4.0 datasets. Additionally, B4 ranks highest in the "behavior but not typical for autism" category, meaning that the behaviors that exist are not normal but also not common among autistic individuals. This also aligns with a lower recall in the GPT-3.5 and GPT-4.0 datasets. Future studies need to explore the interaction of these expert evaluations with precision and recall.

Through the usage of data generated from ChatGPT, we were able to increase the recall performance using single-shot, example-free prompts. This recall increase makes this tool useful for screening potential individuals who may have autism, but the low precision makes it untrustworthy for making an official diagnosis. Curiously, despite there being more A1 samples in the initial dataset than any other criterion, it still had among the largest increases in recall (13% for GPT-3.5 and GPT-4.0 and 19% for GPT-4.0 only, respectively), even more than the much smaller B3 criterion (11% for GPT-3.5 and GPT-4.0 and 14% for GPT-4.0 only, respectively). However, given the issues it faces with precision, this augmentation method still needs more work before it can be fully depended upon as a reliable data augmentation method.

*Limitations*

Our study has several limitations. It should be noted that the data gathered from this study was obtained in September. There is a high possibility that any limitations to OpenAI's GPT models may have changed or no longer exist. Thus, it could be valuable to repeat this study later to measure changes to the study over time. Additionally, this study focused on creating new medical observations, not complete records. More research needs to be conducted to determine if the GPT models could successfully create complete "artificial individuals" that simulate showing those symptoms over a multi-year span such that a medical professional labels the record with the autism diagnosis.

Ideally, the first study should have been completed using only GPT-3.5 data. However, given the limitations of the LLM, it was not possible at this time. As OpenAI updates its ChatGPT models, it could be possible that in a future update, it could handle such a request and may be worth looking into in a future study.

By manipulating gender, we may have reinforced GPT-3.5 and GPT-4.0's potential for bias as it was possible it may create data containing more traditionally masculine or feminine traits despite minimal changes to the prompt. More studies are needed to analyze if such biases exist within the OpenAI's models.

Additionally, while it would have been ideal to compare different Large Language Models, we must also contend with the requirements to run such models. Processing or creating synthetic text data requires massive amounts of computational power. A huge advantage of using ChatGPT is that it requires minimal local computational resources since OpenAI handles all of ChatGPT's backend processing. Ultimately, we do not have the computation resources to run our own versions of their models locally. Future research needs to focus not only on comparing different types of models but also on determining ways to lower their computational costs.

*Potential Implications*

The ability of GPT-3.5 and GPT-4.0 to produce data that resulted in higher recall has value for autism screening models and any other output-sensitive domains, such as screening for medical issues such as cancer, disease, or other ailments. In these instances, while it may occasionally find samples that turn out not to be problematic, these models will be trustworthy if they do rule out the issue it intends to detect.

**Acknowledgment**

This project was supported by grant number R01MH124935 from the National Institute of Mental Health. Part of the data presented were collected by the Centers for Disease Control (CDC) and Prevention Autism and Developmental Disabilities Monitoring (ADDM) Network supported by CDC Cooperative Agreement Number 5UR3/DD000680 and by the University of Arizona FY23 Research, Innovation & Impact (RII) and the Technology Research Initiative Fund/Improving Health Initiative (BIO5 Rapid Grant).

The study was reviewed and approved by the Institutional Review Board at the University of Arizona.**References**
1. Phillips J, Pond D, Goode S. Timely diagnosis of dementia: can we do better. Canberra: Alzheimer's Australia. 2011;
2. Moul JW. Timely diagnosis of testicular cancer. urologic clinics of north america. 2007;34(2):109–17.
3. Bradford A, Kunik ME, Schulz P, Williams SP, Singh H. Missed and delayed diagnosis of dementia in primary care: prevalence and contributing factors. Alzheimer disease and associated disorders. 2009;23(4):306.
4. Dong X, Qian L, Guan Y, Huang L, Yu Q, Yang J. A multiclass classification method based on deep learning for named entity recognition in electronic medical records. In: 2016 New York Scientific Data Summit (NYSDS). 2016. p. 1–10.
5. Shaw KA, Bilder DA, McArthur D, Williams AR, Amoakohene E, Bakian AV, et al. Early identification of autism spectrum disorder among children aged 4 years — Autism and Developmental Disabilities Monitoring Network, 11 Sites, United States, 2020. MMWR Surveill Summ. 2023 Mar 24;72(1):1–15.